\documentclass{article}
\usepackage{graphicx} % Required for inserting images
\usepackage{color}
\usepackage{amsmath}
\usepackage{amsfonts}
\usepackage[top=3cm, bottom=3cm, left=3cm, right=3cm]{geometry}
\usepackage{graphicx}
\usepackage{hhline}
\usepackage{float}
\usepackage{multirow}
\usepackage{array}
\usepackage{soul}
\usepackage{enumitem}

\usepackage[
backend=biber,
style=alphabetic,
sorting=ynt
]{biblatex}
\usepackage[symbol]{footmisc}

\usepackage[T1]{fontenc}
\usepackage{authblk}

\bibliography{tsfm-bib}

\title{Cisco Time Series Model Technical Report}
\usepackage{authblk}

\author{Liang Gou\protect\footnotemark[1]}
\author{Archit Khare\protect\footnotemark[1]}
\author{Praneet Pabolu\protect\footnotemark[1]}
\author{Prachi Patel\protect\footnotemark[1]}
\author{Joseph Ross\protect\footnotemark[1]}
\author{Hercy Shen\protect\footnotemark[1] \protect\footnotemark[3]}
\author{Yuhan (Ellen) Song\protect\footnotemark[1]}
\author{Jingze Sun\protect\footnotemark[1]}
\author{Kristal Curtis\protect\footnotemark[2]}
\author{Vedant Dharnidharka\protect\footnotemark[2]}
\author{Abhinav Mathur\protect\footnotemark[2]}
\author{Hao Yang\protect\footnotemark[2]}

\affil{\normalfont{\texttt{\{lgou, arckhare, ppabolu, prachpat, josross, hershen, elsong, jingzsun, krcurtis, vdharnid, abmathu2, haoya2\}@cisco.com}}}

\date{\today}

\begin{document}
\newcommand{\checkme}[2]{{\color{red}{\textbf{CHECK [{#1}]: }{#2}}}}
\newcommand{\todo}[2]{{\color{red}{\textbf{TODO [{#1}]: }{#2}}}}
\newcommand{\praneet}[1]{{\color{blue}{\textbf{PP: }{#1}}}}
\newcommand{\liang}[1]{{\color{green}{\textbf{LG: }{#1}}}}
\newcommand{\kristal}[1]{{\color{cyan}{\textbf{KC: }{#1}}}}

\maketitle
\footnotetext[1]{These authors contributed equally to the core development of this work, listed alphabetically by last name.}
\footnotetext[2]{These authors contributed equally to supporting and extending this work, listed alphabetically by last name.}
\footnotetext[3]{Hercy Shen contributed to this work while an intern at Splunk.}

\begin{abstract}

We introduce the Cisco Time Series Model, a univariate zero-shot forecaster. 
This time series foundation model is the result of a general architectural innovation to a time series model enabling it to accept multiresolution input, applied to a popular decoder-only time series model (TimesFM). The resulting multiresolution decoder-only model is trained on over 300B unique data points, with more than half coming from the observability domain.
Quantitative and qualitative evaluations demonstrate that the resulting model achieves superior performance on observability datasets while retaining very similar performance on a standard general-purpose forecasting benchmark (GIFT-Eval), and suggest that the multiresolution structure enables the model to make more accurate predictions on long context input.

\end{abstract}

\section{Introduction}
\subsection{Background}
Modern LLMs are capable of learning complex statistical properties of language from a vast corpus of text. Rather than being trained to emulate a particular style or perform a particular task, they learn structure across diverse examples of token sequences, and the learned representations can be transferred to many downstream tasks and applications. The main idea of a time series foundation model (TSFM) is to apply the same playbook -- including the  transformer architecture that has revolutionized natural language processing -- to sequences of numerical data, i.e., time series.
%There are now several approaches to TSFM utilizing the. 
Our present focus is to train a univariate TSFM capable of high-quality zero-shot forecasting, with emphasis on time series arising in certain business domains (initially, observability).
Thus, having been exposed to patterns across many time series during training, given a segment of a new (unseen) time series, the TSFM is expected to predict its subsequent segment without any auxiliary parameter adjustment or fitting.

Architectural differences among TSFMs can be found in their approaches to tokenization, transformer configuration, and prediction heads. PatchTST \cite{PatchTST} introduces the idea of a time series patch as the analogue of a token, uses a linear transformation of a patch as a replacement for the token embedding, and finally applies a standard transformer encoder architecture. TimesFM \cite{TimesFM} uses a residual block to embed time series patches, enabling learning of more complex representations, and applies a decoder-only architecture. Chronos \cite{chronos} tokenizes individual data points via scaling and then applies the (encoder-decoder) T5 architecture \cite{raffel2020exploring}, notably formulating forecasting as a classification problem; subsequent versions (Chronos-Bolt, Chronos-2 \cite{chronos2}) utilize patching and ``meta features'' before applying transformer layers, and Chronos-2 uses a T5 encoder.
Moirai \cite{moirai} utilizes multiple patch sizes and also learned a mixture of distributions, elevating probabilistic forecasting to a first class consideration; Moirai-MoE \cite{moirai-moe} applies the mixture of experts pattern as a learnable replacement for various frequency heuristics.
Toto \cite{toto}, \cite{toto-2} uses a causal patching mechanism, a learned mixture of $t$-distributions for the prediction head,  and a composite loss function; its training corpus and accompanying BOOM benchmark are based heavily on observability data.

There have been several efforts to incorporate multiscale structure in TSFMs.
Pyraformer \cite{pyraformer} uses convolutions at multiple scales to build a multiresolution representation, then applies attention in a pyramidal pattern to share information across resolutions.
Scaleformer \cite{scaleformer} processes the same input at multiple scales, proceeding from coarser to finer, using average pooling and upsampling to translate across resolutions. 
Pathformer \cite{pathformer} introduces an adaptive multi-scale transformer block, giving the patch size a dynamic flavor; it also intentionally models trend and seasonality. %\todo{JOE}{others}
Multi-resolution time series transformer \cite{zhang2024multi} iteratively applies attention directly to several patchings of the same time series (with different length and stride); the attention operates separately on each patching, and the results are combined.

{While existing TSFMs have introduced a variety of architectural innovations—from patch-based tokenization to mixture of experts designs—the majority remain constrained by relatively limited context windows, typically spanning 512 to 4,096 data points, with the latest TimesFM 2.5 \cite{TimesFM} extending this to 16,384. 
All of the more complex multiresolution architectures mentioned above share the characteristic that they process the same input at multiple resolutions, so have no particular suitability for long context.
This limitation hampers their ability to effectively leverage long historical sequences, which are often critical for accurate forecasting in domains (such as observability)  where past patterns persist over extended periods. Addressing this gap is central to our work: we propose an architecture and data handling methodology that explicitly target improved modeling in long context scenarios.}
In contrast to prior multiresolution approaches, we view the coarser resolution context as a potential asset to make the finer resolution predictions more accurate.
A very practical motivation for our architecture is that time series data is often available at different resolutions according to age (fine resolution data ``expires'' and is aggregated into coarse resolution summaries), and 1-minute and 1-hour resolutions in particular are often persisted. 
Our model is able to exploit pre-computed rollups in  scenarios where full history at the finest resolution may not be available.
We achieve efficient use of long context and a better tradeoff between recent detail and historical context as our model operates directly on multiresolution input: the more complex multiresolution architectures would require a context length of 30,720 (30 times as long as ours) 
to cover the same time range.

\subsection{Contributions}
In the realms of theory and methodology, we describe 
a novel multiresolution architectural pattern applicable to most TSFMs, and 
principles for data operations (filtering, sampling, etc.) on time series especially relevant for adapting a TSFM to a new domain.
To realize and evaluate these data and modeling principles, we report the performance of the Cisco Time Series Model. This work makes three main contributions to time series foundation modeling.

\textbf{Multiresolution TSFM Architecture.} We propose a novel multiresolution architectural pattern explicitly designed to leverage long‑context information. The architecture balances global temporal patterns obtained from coarse, low‑resolution views with fine‑grained temporal details derived from high‑resolution sequences, enabling accurate and timely forecasting across diverse time series domains.

\textbf{Principles for Time Series Data Operations.} We formalize practical techniques for filtering, sampling, and resolution handling tailored to TSFMs. These principles streamline adaptation to new domains by ensuring data consistency, preserving key temporal characteristics, and optimizing computational efficiency during training and inference.

\textbf{Comprehensive Evaluation and Qualitative Case Studies.} We implement the proposed architecture and data operation principles in the Cisco Time Series Model, trained on a curated dataset aligned with our methodology. Quantitative evaluation on the GIFT‑Eval benchmark shows that our model matches or exceeds the performance of its pretrained base, with measurable gains in long‑context scenarios. A domain‑specific comparison on observability time series from Splunk’s own usage of the Splunk Observability Cloud demonstrates marked improvements over competing TSFMs. We complement the metrics with qualitative case studies, highlighting examples where the model excels in long‑context forecasting and identifying short-context cases where challenges remain.

Overall, our findings indicate that the proposed multiresolution architecture and data operation framework significantly enhance TSFM performance in the observability domain while maintaining robust general time series forecasting capabilities. Model weights \cite{huggingface} and inference code \cite{github} are provided under an Apache 2.0 license.

\section{Model}

\subsection{Motivation}
The size of the context window of a TSFM typically ranges from 512 to 4,096 data points. 
For 1-minute resolution data, this suffices to capture intraday patterns and rudimentary diurnal structures, but is not nearly long enough to see week-over-week periodicity and growth/decay. Consider the following time series in Figure \ref{fig:first_sawtooth}, with both 512 hours (purple) and 512 minutes (green) of history displayed:

\begin{figure}[h!]  % h=here, t=top, b=bottom, p=page
    \centering
    \includegraphics[width=0.7\textwidth]{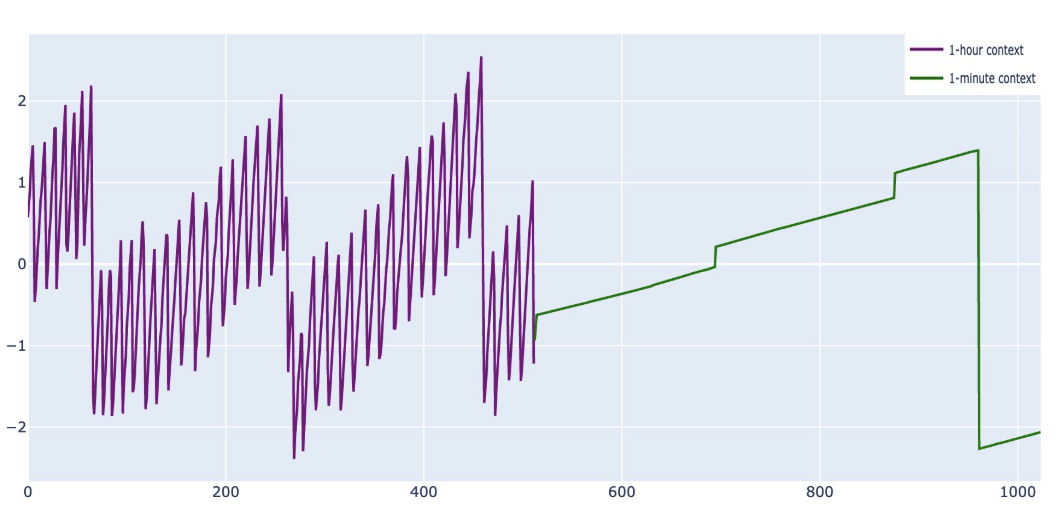} % path to your image
    \caption{Time series with structure in 1-hour resolution not visible at 1-minute resolution}
    \label{fig:first_sawtooth}
\end{figure}

\begin{figure}[h!]
    \centering
    \includegraphics[width=0.9\textwidth]{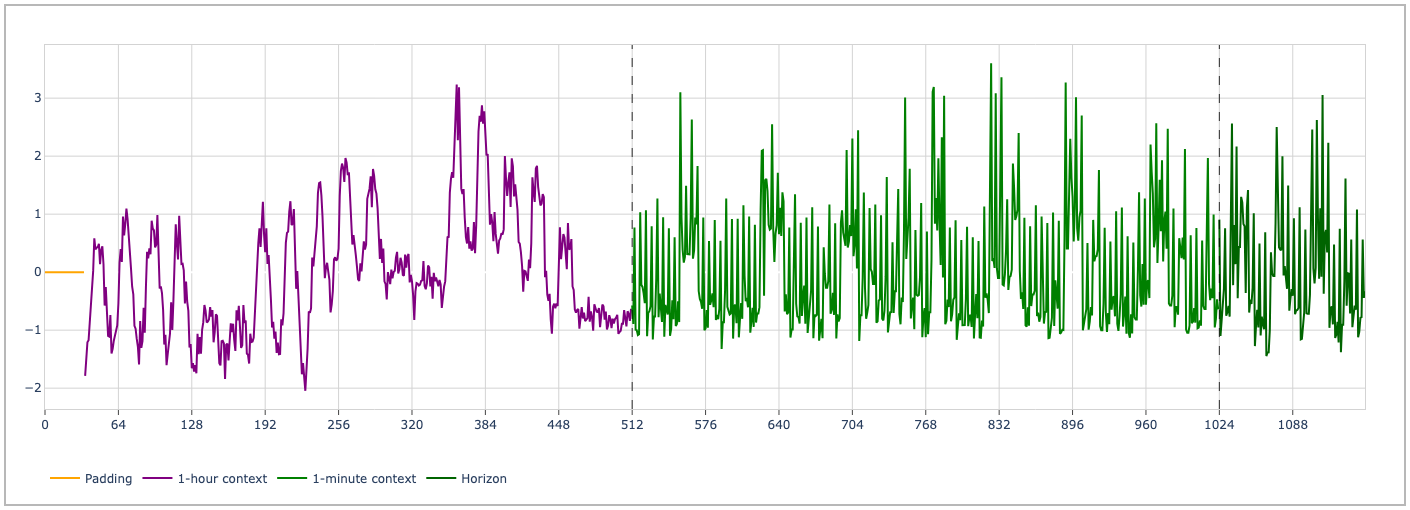}
    \caption{Multiresolution time series with padded 1-hour context %\kristal{not sure if we have time to fix this but IMO the light and dark greens are hard to distinguish so not that obvious where the horizon is}
    }
    \label{fig:multi_resolution_time_series_example_v2}
\end{figure}

From the 512-minute history, no patterns are evident; while 512 hours appears sufficient, local detail and timeliness are compromised with forecasts or detections computed on 1-hour data. 
Our core modeling philosophy is that it is possible to fuse the information present at different timescales/resolutions, and this leads to superior outcomes in real operational scenarios. Figure \ref{fig:multi_resolution_time_series_example_v2} illustrates the multiresolution context and the (fine resolution) horizon of an example time series with less than 512 hours of data (hence a padded 1-hour context).

\subsection{Architecture} \label{model:arch}
\subsubsection{Overview}
Forecasting models generally operate at a single resolution, using context (input) at 1-minute resolution to predict the horizon (output) at 1-minute resolution, context at 1-hour resolution to predict the horizon at 1-hour resolution, and so on.
We propose a \textit{multiresolution} architecture to enable learning time series structures (trends, seasonality, etc.)~at different scales, and synthesizing these learned structures into better forecasts. 
We aim to learn a function $F$ mapping a multiresolution context, e.g., a pair
(512 1-hour data points, 512 1-minute data points), to a horizon at the finer resolution, e.g., 128 1-minute data points. Here, the 1-hour portion of the context represents the 512 hours preceding the horizon, and similarly, the 1-minute portion represents the 512 minutes preceding the horizon. The ratio of the resolutions should be the same for all inputs; we fix the ratio at 60.

We utilize two auxiliary structures to allow the model to more cleanly distinguish the two contexts: a \textbf{special token} (analogous to special tokens in language modeling) between the coarse and fine resolution series with a learnable representation, and learnable \textbf{resolution embeddings} which are added to the input prior to the transformer layers. The special token operates in ``sequence space'' while the resolution embedding operates in ``model space.'' The resolution embedding is conceptually similar to the frequency embedding of \cite{TimesFM}. Ablation studies for these structures are summarized in Section \ref{sec:ablation}.

\begin{figure}[h!]
    \centering
    \includegraphics[width=1\linewidth]{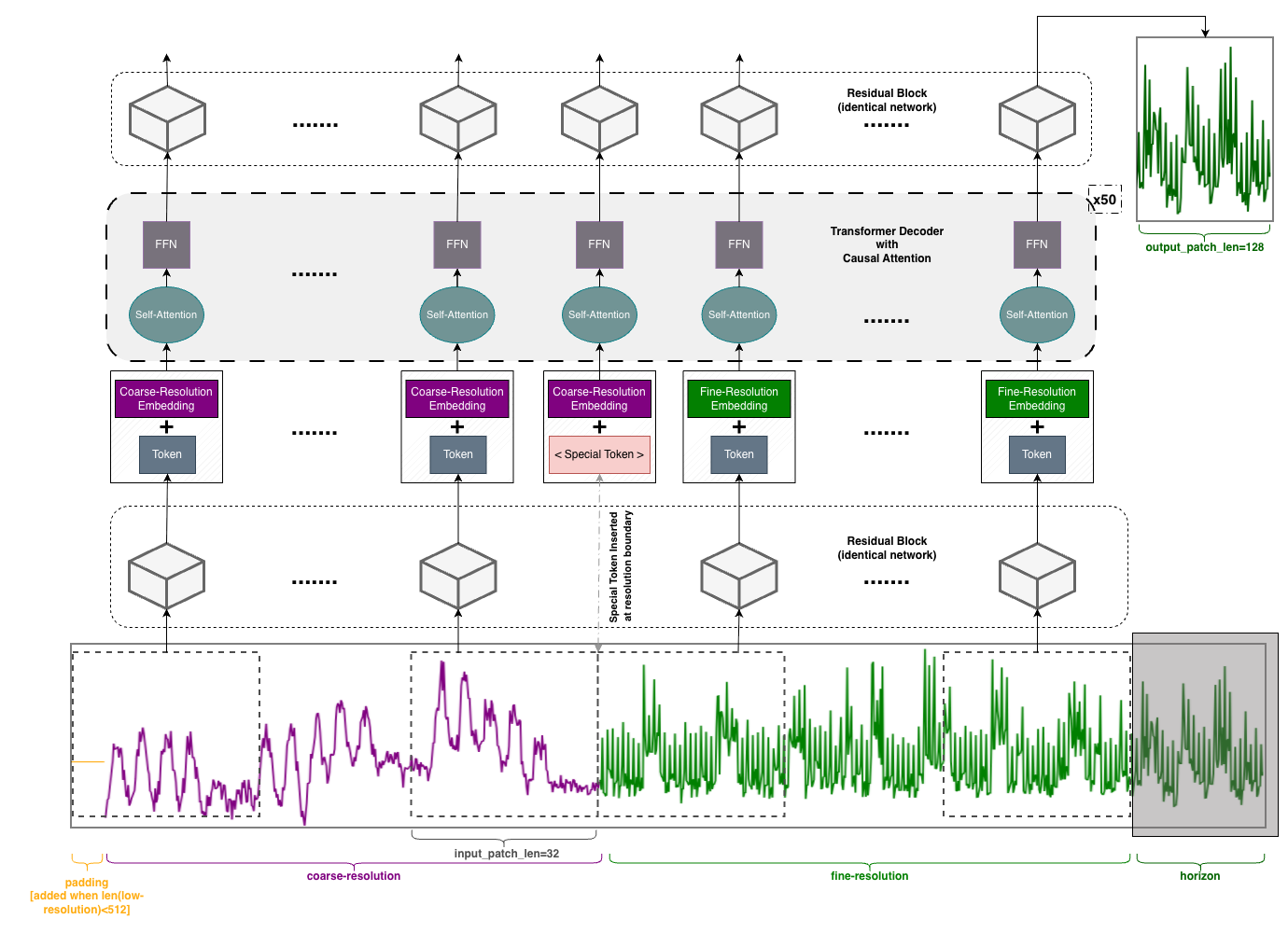}
    \caption{
    Architecture diagram illustrating Resolution Embeddings and Special Token.}
    \label{fig:tfm_mr_resemb_spec_token}
\end{figure}

The coarse resolution context precedes the fine resolution context, and no positional encoding is used. The rest of our architectural choices, and indeed our implementation, follows TimesFM \cite{TimesFM}: $F$ is composed of patching and padding, a residual block for patch embedding, decoder-only transformer layers, and another residual block for un-embedding. The next section provides a detailed description with an emphasis on our new components. The overall architecture is depicted in Figure \ref{fig:tfm_mr_resemb_spec_token}; we suspect the same multiresolution pattern can be applied to many TSFMs.

\subsubsection{Model Input and Output}

The coarse and fine contexts are standard normalized independently, and the statistics of the finer context are used to normalize the horizon. The objective is then to compute the normalized horizon as a function of the two normalized contexts. At inference time, the normalization is undone to produce a forecast.

\textbf{Context normalization.} Let $(x_c, x_f) \in \mathbb{R}^{512} \times \mathbb{R}^{512}$ denote a multiresolution context, and $y \in \mathbb{R}^{128}$ the corresponding horizon. Let $\mu_c, \sigma_c$ and $\mu_f, \sigma_f$  represent the mean and standard deviation of the first 32 points of $x_c$ and $x_f$, respectively. (Typically $\mu_c$ will be comparable to $\mu_f$, but $\sigma_c$ and $\sigma_f$ may be quite different depending on the dynamics.) This yields normalized inputs $x_c \leftarrow \frac{x_c - \mu_c}{\sigma_c}$ and $x_f \leftarrow \frac{x_f - \mu_f}{\sigma_f}$.

\textbf{Patching.} Following the process described in TimesFM \cite{TimesFM}, the normalized inputs are decomposed into contiguous non-overlapping patches of length $\texttt{input\_patch\_len}=32$, along with a mask for each patch.

\textbf{Tokenization.} Writing $x_c = (u_1, \ldots, u_{16})$ and $x_f=(u_{17}, \ldots, u_{32})$, with each patch $u_i \in \mathbb{R}^{32}$, we apply a residual block $g_{\text{in}} : \mathbb{R}^{32} \to \mathbb{R}^{1280}$ to each patch: $g_{\text{in}}(u) = W_o\phi(W_h u)+W_r u$, where $W_0, W_h, W_r$ are linear layers and $\phi$ is a SiLU activation. This yields tokens $h_i = g_{\text{in}}(u_i) \in \mathbb{R}^{1280}$.

We then add our new architectural changes as follows:
\begin{itemize}
    \item \textbf{Special Token (ST).} This is inserted between the resolution token streams, yielding $$[h_1, \ldots, h_{16}, ST, h_{17}, \ldots, h_{32}].$$ We insert a corresponding zero in the padding mask.
    % $\tilde{x}=[x^{\text{c}},\,ST,\,x^{\text{f}}]$ along with a padding mask 
    % $p=[p^{\text{c}},\,0,\,p^{\text{f}}]$ and a keep-mask that drops $ST$ after decoding.
    \item \textbf{Resolution Embeddings (RE).} Each token $h_i$ is incremented by $r_i=\mathrm{RE}(z) \in \mathbb{R}^{1280}$ with $z=0$ for fine resolution tokens (and $ST$) and $z=1$ for coarse resolution tokens, i.e., $h_i \leftarrow h_i + r_i$.
    
    % $r_n\in\mathbb{R}^D$ and all these tokens are updated as: \[x\_{n}\leftarrow x\_{n}+r\_{n}\]
\end{itemize}

\textbf{Outputs.} The resulting tokens are processed by 50 decoder-only transformer layers and another residual block mapping to the length of the forecast horizon (i.e., $g_{\text{out}} : \mathbb{R}^{1280} \to \mathbb{R}^{128}$). Overall, this procedure describes a function $F : \mathbb{R}^{512} \times \mathbb{R}^{512} \to \mathbb{R}^{128}$ operating on a normalized input. Then, the objective of training is to learn a function $F : \mathbb{R}^{512} \times \mathbb{R}^{512} \to \mathbb{R}^{128}$ such that $F( \frac{x_c - \mu_c}{\sigma_c}, \frac{x_f - \mu_f}{\sigma_f})$ is close to $\frac{y - \mu_f}{\sigma_f}$. At inference time, the next 128 steps of the context $(x_c, x_f)$ are forecasted to be $\mu_f + \sigma_f  F( \frac{x_c - \mu_c}{\sigma_c}, \frac{x_f - \mu_f}{\sigma_f})$.

\textbf{Autoregressive multiresolution decoding.} 
Given a multiresolution context $(x_c, x_f)$, a decode step generates mean and quantile forecasts for $L = \texttt{output\_patch\_len}=128$ fine resolution points. We write $K =\texttt{resolution\_ratio}$; for us, $K=60$. We append the mean forecasts to the fine resolution context and zeros to the corresponding padding mask. To update the coarse resolution context, we aggregate the first $K \lfloor L / K \rfloor$ predictions in the fine resolution forecast: we append 
$$[ \frac{1}{K}\!\sum_{j=1}^{K}\hat{y}_{(k-1)K+j}\,]_{k=1}^{\lfloor L/K\rfloor}$$
\noindent 
to the coarse resolution context.
These concatenated patch outputs form the final returned mean and quantile forecasts.

\subsection{Training}

Our architecture is a modification of the TimesFM \cite{TimesFM} architecture, adding special token and resolution embeddings, with 500M parameters in total. We experimented with various freezing/un-freezing schedules (e.g., focusing on the new components first, gradual thawing) but found continued pre-training to yield the fastest convergence, including when compared to training from scratch.

Biases, norms, and embeddings are optimized via AdamW, while hidden layers are optimized via Muon. We use a learning rate of 5e-5 for the AdamW parameters and 1e-4 for the Muon parameters, and both learning rates decay according to a Cosine Annealing schedule. The batch size is set to 1024 per core. We also use a loss clipping %of 25.0 
that roughly corresponds to $(5\sigma)^2$ in normalized space, a weight decay of $0.05$, and gradient clipping. % of 1.0. 
The loss function combines both the mean squared error of the point predictions and the quantile loss of the quantile predictions at $q=0.1, 0.2, \ldots , 0.9$. We train for 20 epochs and choose the checkpoint with smallest validation loss; in most experiments, the best epoch is between 5 and 10.

Small-scale experiments were performed on a single node, but the larger-scale training used to produce our published checkpoint utilized 64 Nvidia H200 GPUs with Distributed Data Parallel (DDP) training on 8 nodes, leading to an effective batch size of 65,536. To account for this large batch size, we scaled the learning rate by the square root of the number of nodes (i.e., $\sqrt{8}$). Each epoch in our large-scale training took about 30 hours to complete at full precision.\\

\subsection{Future Generalizations}
The fixed placement of the token is perhaps too rigid, 
and can be seen as an inefficient compression of certain inputs: if one wanted a forecast given 24 hours of historical data, for example, one might ask for a (1-hour, 1-minute) context window structured as $((x_{c, 1}, \ldots , x_{c, 24}), (x_{f, 1}, \ldots , x_{f, 1000})) $ rather than $((x_{c, 1}, \ldots , x_{c, 512}), (x_{f, 1}, \ldots , x_{f, 512})) $, where most of $x_c$ is padded.
In a similar direction, to fuse information across more than two resolutions, it might be interesting to use multiple special tokens and resolution embeddings.

\section{Data}
\label{sec:data}

\subsection{Sources}

\textit{Observability} refers to the practice of turning telemetry (logs, metrics, traces, events) into a tool for managing the health and performance of digital applications. The primary novel data source we bring to TSFM is a vast collection of \textbf{proprietary} metric time series from the {observability} domain, namely Splunk's own usage of the metrics subsystem of the Splunk Observability Cloud. Examples of time series include: infrastructure metrics (utilization of core resources such as CPU, memory, disk), application or service metrics (request rates, errors, latency), real-user metrics (measured in the browser or mobile device), and metrics for specific technology components (e.g., queues, key-value stores, analytics engines). Altogether, we have collected approximately 400M time series across a period of 13 months. The series are collected at 1-minute resolution. For additional variety, some series are aggregated to a 5-minute resolution.

The \textbf{public} data sources utilized are the GIFT-Eval pretraining corpus \cite {GIFT} and the Chronos datasets \cite{chronos} (excluding the GIFT-Eval test set).
To get a rough sense of the scale and diversity, GIFT-Eval consists of 4.5M time series and 230B data points in total, drawn from seven domains (web, econ/finance, energy, etc.); Chronos consists of almost 900K time series and 85B data points in total.
%\kristal{elsewhere we say "B" instead of spelling out billion; should we stick with one or the other?}
We also leverage \textbf{synthetic} data, utilizing the KernelSynth method \cite{chronos}. Since we target relatively high-resolution operational scenarios, our periodic kernels use smaller periodicity parameters.

\subsection{Operations}

\subsubsection{Basic Filtering and Transformation}
Some heuristic filtering is performed at the time series level: stretches with too many missing values are ignored, as we do not want to train or evaluate on heavily imputed data. We then use last value extrapolation to fill in any remaining (short stretches) of missing values.
Various types of ``sparse'' time series often arise from events being modeled as time series, and do not really belong to the time series domain.

The basic metric types found in observability are gauges, (possibly distributed) counters, and cumulative counters. We found that including too many raw cumulative counters in a training set gave the resulting model a ``default linear extrapolation'' character, and so applied the difference operator to these time series.

\subsubsection{Preparation for Modeling}
A single time series is typically much longer than the context window size. Via a sliding window construction, a single time series produces many context-horizon pairs. Train/validation/test splits of context-horizon pairs are both temporally-aware and time series identity-aware. More specifically, the collection of windows derived from a time series is split in temporal order into train, validation, and test examples. The temporal split ensures no leakage from the future into the past and enables \textit{in-domain} evaluation appropriate for statistical models that require fitting to some history of a time series before generating predictions. Those time series not part of any training or validation set are considered \textit{out-of-domain}, often called zero-shot in the foundation model setting.
Since we are aware of correlations across time series, especially within the observability domain, we also globally segregated test data across time: 
this means evaluating on a segment of a time series where no part of the time series was seen during training, and every timestamp in the test data is subsequent to the latest timestamp in the train and validation sets used to fit and select the model. These test data are both out-of-domain and in-the-future.

\subsubsection{Filtering and Sampling} \label{sec:sampling}
At the window level, more advanced filtering and downsampling is applied as well. These rules reflect our beliefs about what is learnable, and about how time series forecasts should be evaluated. These rules are applied to the fine resolution windows.
\begin{itemize}
\item We filter out windows with excessive ``flat spots” (segments with no change that are too long) and also too few unique values; in the observability domain, these are often event data masquerading as time series.

\item We compute the maximum absolute deviation from the median on both the context and horizon, and apply a threshold to the ratio of the value on the horizon to that on the context. The underlying heuristic is that we do not expect any model to be able to learn step changes in the horizon from a stationary context, nor does it make sense to evaluate any method by its inability to predict a step change.
    
\item Windows with large spectral entropy are downsampled to arrange a ``healthy'' distribution of entropy. The motivation is that time series drawn i.i.d.~(from any distribution) have large entropy 
and no learnable structure; for such series the model should not attempt to learn anything beyond basic statistics of the series, and there is little to evaluate.

\end{itemize}

\subsubsection{Diversity}
\label{sec:diversity}

We implemented multiple data controls to encourage diversity in the training data. 

\begin{figure}[H]  % h=here, t=top, b=bottom, p=page
    \centering
    \includegraphics[width=0.7\textwidth]{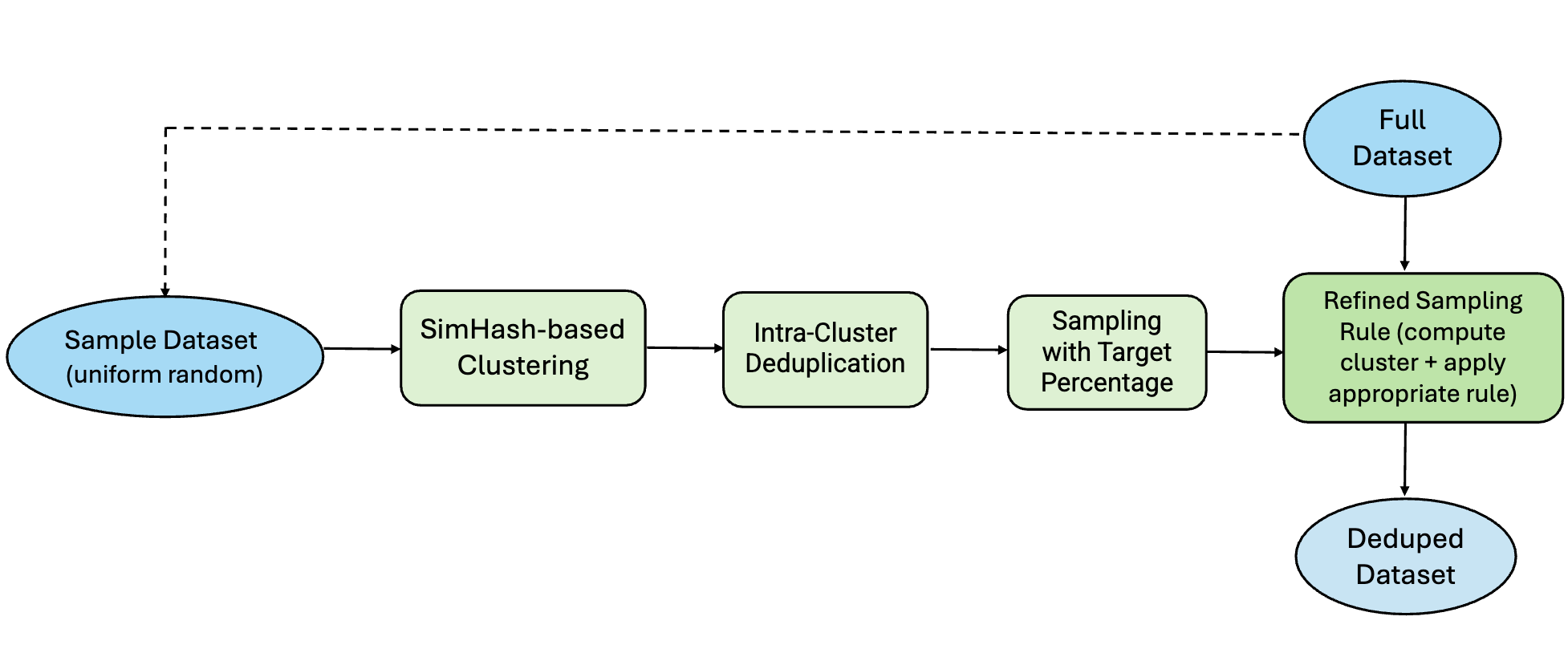} % path to your image
    \caption{Statistical deduplication pipeline}
    \label{fig:dedupe_pipeline}
\end{figure}

\begin{enumerate}
\item \textbf{Raw diversity in data sources.} 
The mixture of the different data sources is a primary lever in ensuring diversity in data and good generalization properties of the trained model.
Adding even a small amount of a new data source tends to make the model better, and there is a large stable range where refined tradeoffs (e.g., modest improvement in one domain in exchange for slight deterioration in a few other domains) can be achieved by altering the data proportions.
Within GIFT-Eval, to prevent the larger datasets from dominating the modeling datasets, the stride length in the sliding window construction is adapted to the dataset size.

\item \textbf{Padding awareness.}
Experiments indicate the model’s performance across test examples with different padding amounts is sensitive to the padding mix seen during training. To ensure good performance across different padding amounts, we arrange for the training data to have good representation across different padding amounts. This can be seen as an analogue of training on partially occluded images, to ensure, for example, that an image classifier does not rely too heavily on any particular feature of a class. 

\item \textbf{Statistical deduplication.} A coarse clustering is obtained via the locality sensitive hashing (LSH)
technique SimHash \cite[Sec.~3]{charikar2002similarity}:
this constructs a binary code for each time series, and the code can be used as a cluster label. We aim to roughly equalize the contributions of the largest 50\% of clusters. To reduce the size of the larger clusters, we use a sample to compute both a medoid and a histogram of distances from this medoid (for each cluster). When the distribution of distances has a spike near zero, we infer that the same pattern has many near-facsimiles and use a distance-based sampling rule to encourage diversity within the cluster. When the distribution of distances is ``healthy,'' we perform random sampling from the cluster. The overall two-stage sampling rule (apply SimHash, apply appropriate sampling rule) can be applied to the entire dataset with no coordination among data shards. 
This operation is applied to the fine resolution windows.
Figure \ref{fig:dedupe_pipeline} illustrates the overall procedure. 
\end{enumerate}

\begin{figure}[htbp]
    \centering
    % First image
    \begin{minipage}{0.45\textwidth}
        \centering
        \includegraphics[width=\textwidth]{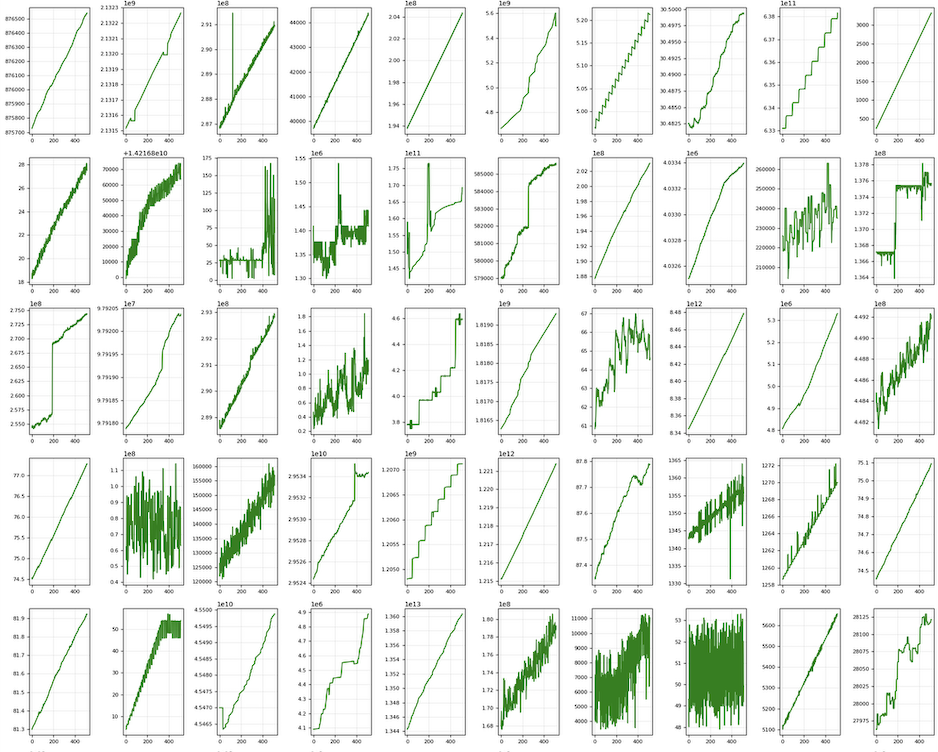}
        \caption{50 random samples from the largest cluster before deduplication
        }
        \label{fig:before_dedupe}
    \end{minipage}
    \hfill
    % Second image
    \begin{minipage}{0.45\textwidth}
        \centering
        \includegraphics[width=\textwidth]{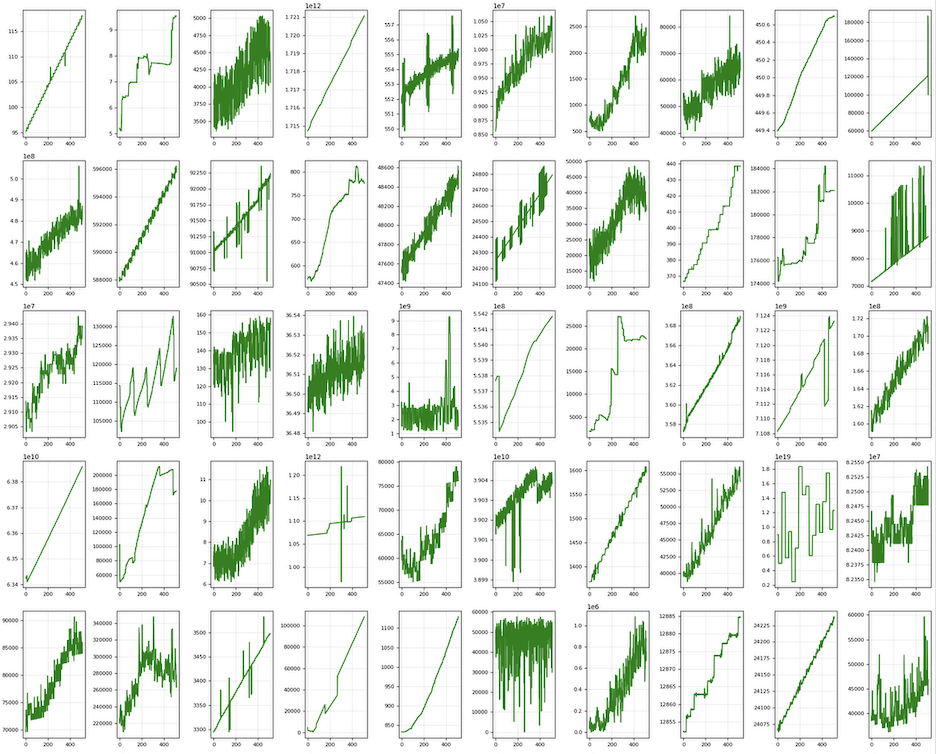}
        \caption{50 random samples from the largest cluster after deduplication
        }
        \label{fig:after_dedupe}
    \end{minipage}
\end{figure}

\begin{figure}[H]  % h=here, t=top, b=bottom, p=page
    \centering
    \includegraphics[width=0.99\textwidth]{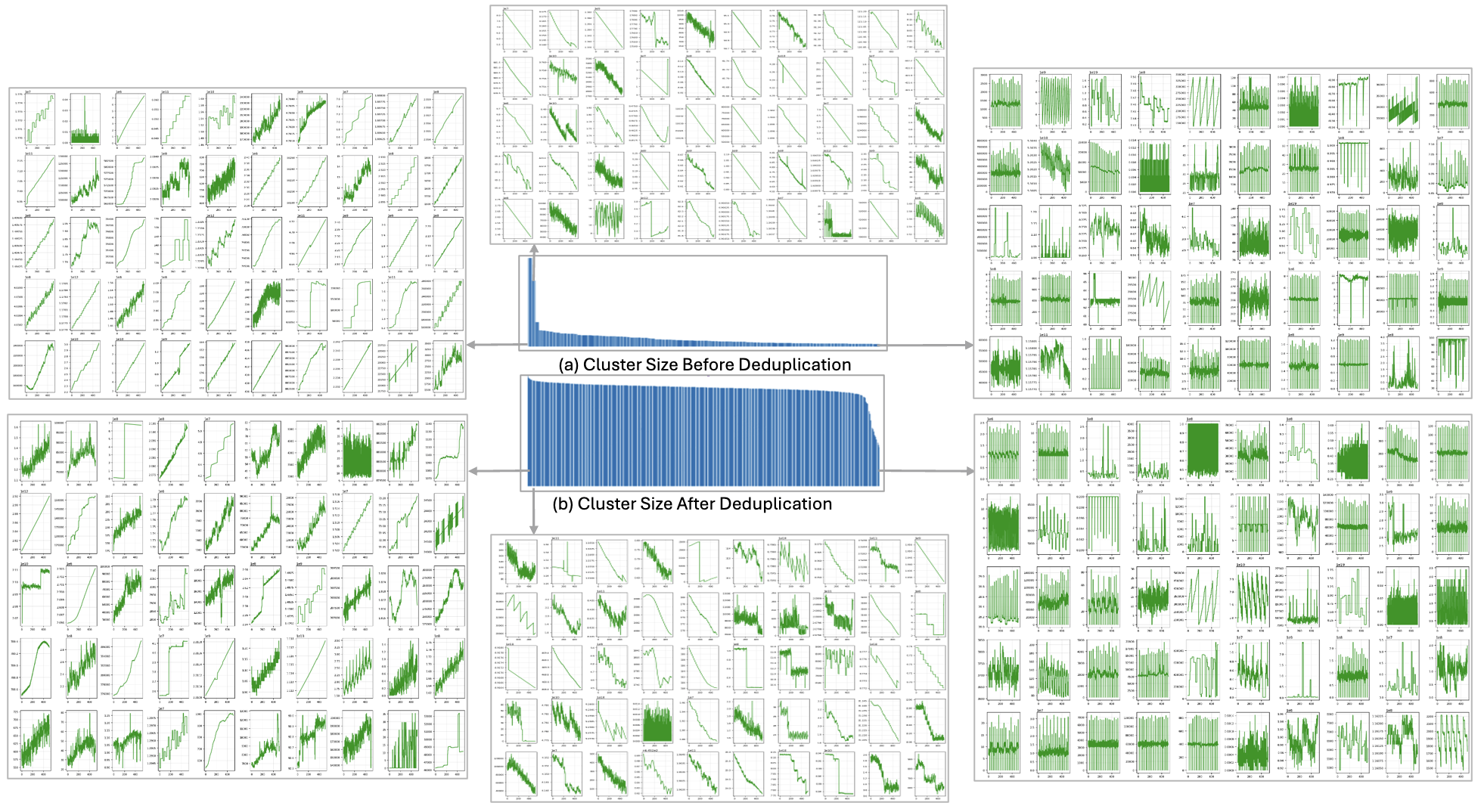} % path to your image
    \caption{Examples from the largest, second largest and smallest cluster (left to right) before and after deduplication; histogram of cluster sizes before and after deduplication.}
    \label{fig:three_clusters}
\end{figure}

\begin{figure}[h!]  % h=here, t=top, b=bottom, p=page
    \centering
    \includegraphics[width=0.9\textwidth]{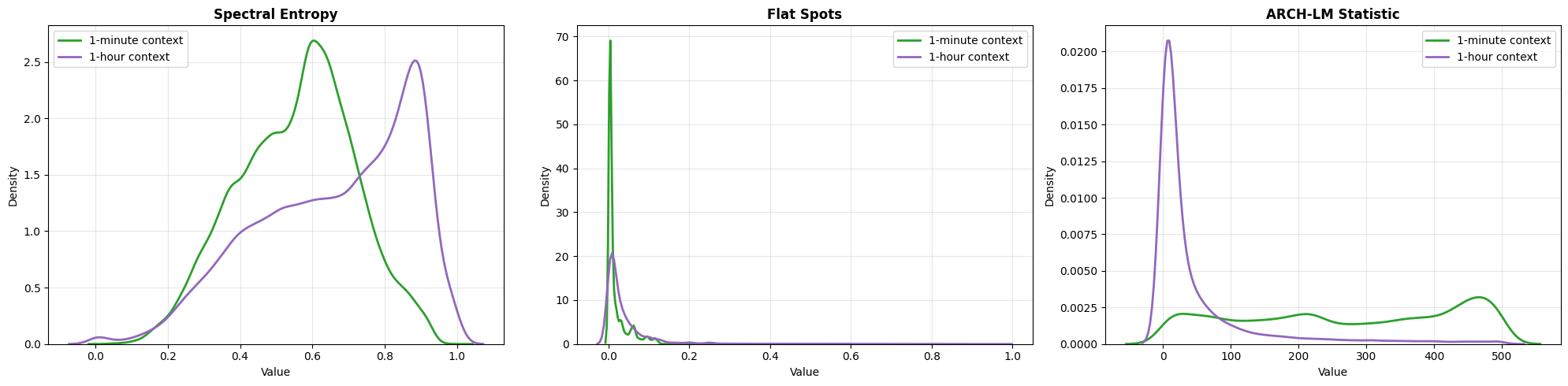} % path to your image
    \caption{Density distributions of three statistical measures for both 1-hour and 1-minute context, observability dataset. The flat spots value is the normalized length of the longest flat spot.}
    \label{fig:ts_stats}
\end{figure}

Figure \ref{fig:before_dedupe} shows 50 random time series windows from the largest cluster prior to the deduplication procedure. There are many near-repetitions of the same simple pattern. Figure \ref{fig:after_dedupe} shows a sample from the same cluster after deduplication, where the variety is much richer. Figure \ref{fig:three_clusters} illustrates the overall effect of deduplication. 
See Figure \ref{fig:ts_stats} for a statistical profile of the resulting observability dataset. The 1-minute contexts tend to have lower spectral entropy and fewer flat spots due to the sampling rules as described in Section \ref{sec:sampling}, and also tend to have more volatility clustering.


Overall, these filtering, sampling, and deduplication rules can be seen as rough time series analogues of cleaning text from the web, e.g., \cite{raffel2020exploring}.  In total, our training set consists of over 300B unique data points, according to the following rough proportions:

\begin{itemize}[noitemsep]
    \item 35\% 1-minute resolution observability,
    \item 16.5\% 5-minute resolution observability, 
    \item 29.5\% GIFT-Eval pre-training corpus,
    \item 4.5\% Chronos pre-training corpus, and
    \item 14.5\% synthetic.
\end{itemize}

\section{Evaluation}

\subsection{Methodology}

Since the TSFMs we compare to operate at a single resolution, we run comparisons where we provide those models only the (length 512) fine resolution context, and another set of comparisons where we provide those models
1024 data points at the finer resolution; in the latter case, note that while the overall length of the context consumed is the same, the models are actually seeing different contexts. We use the notation \textbf{Model (context length)} in the tables. Please note the evaluation datasets for length 512 and length 1024 fine resolution context are different.
In all comparison tables, the models are being asked to predict exactly the same horizons.

The AutoARIMA baseline, not being a zero-shot forecaster, is permitted to use the input context as follows: 
an automated model selection procedure is used to find the parameters $(p, d, q)$ best fitting the context (whether 512 or 1024 data points); the resulting model is then used to forecast the subsequent 128 data points.
This means AutoARIMA adapts its structure to the observed data, whereas the foundation models process the input context without any parameter updates.

\subsection{Metrics}

We use the median prediction (i.e., $q=0.5$) for all models. The basic forecasting error metrics are mean squared error (MSE) and mean absolute error (MAE). MAE is comparatively less influenced by outliers. We also report two normalized variants of MAE: mean absolute scaled error (MASE), a variant normalized by the error of a naive last-(seasonal-)value prediction; and symmetric mean absolute percentage error (sMAPE), a variant normalized by the sum of the absolute values of the actual and predicted values. Note that these normalizations occur at the horizon level. We also report the Mean Scaled Interval Score (MSIS), which is an interval version of the MASE, with $\alpha = 0.05$. 
The continuous ranked probability score (CRPS) is a weighted average of the quantile losses.
The aggregation details are discussed within Sections \ref{o11y_data} and \ref{gift_data}.

\subsection{Results}

\subsubsection{Observability Data} \label{o11y_data}
With reference to the training and validation sets, these time series are both out-of-domain and in-the-future. We apply curation rules similar to those described in Section \ref{sec:data}
to ensure a diverse and high-quality benchmark. 

We report metrics on both 1-minute and 5-minute resolution data (with coarse contexts at 1-hour and 5-hour resolution, respectively). For observability data, no natural seasonality is available, so we present also a Naive baseline which simply forecasts the final value in the context for all time steps in the horizon.
Error metrics are computed per horizon, then all horizons are aggregated via arithmetic mean. This quantity is then normalized by a similar computation using the Naive baseline.
The Naive baseline is not itself normalized by another model; its errors are computed in the context-normalized horizon.

\begin{center}
\begin{table}[ht]% Try here, and then top
\small
%\hhline{|=|=|=|=|=|=|=|}
\centering
\renewcommand{\arraystretch}{1.2}
\begin{tabular}{ |l|c|c|c|c|c|c| } 
 \hline
  & MSE & MAE & MASE & sMAPE & MSIS & CPRS \\ 
\hhline{|=|=|=|=|=|=|=|}
 Cisco TSM (512, 512) & \textbf{0.8524} & \textbf{0.4788} & \textbf{0.4569} & 0.7758 & \textbf{0.1207} & \textbf{0.4126} \\ 
%\hhline{|=|=|=|=|=|=|=|}

\hline

 TimesFM-2.5 (512) & 0.8838 & 0.6265 & 0.7290 & 0.8297 & 0.1732 & 0.5089  \\ 
\hline
 TimesFM-2.0 (512) & 0.8871 & 0.6315 & 0.6722 & 0.9379 & 0.1844 & 0.5467 \\ 

\hline
 Chronos-2 (512) & 0.8816 & 0.6023 & 0.7056 & 0.7811 & 0.1773 &  0.4878\\ 

 \hline
 Chronos-Bolt (512) & 0.9154 & 0.7575& 0.8079 & 0.9755 & 0.2965 & 0.6488 \\ 
\hline
 Toto-1.0 (512) & 0.8836 & 0.6055 & 0.6834& \textbf {0.7741} & 0.2032 & 0.4932\\ 
\hline
 AutoARIMA (512) & 4.0520 &0.8545 & 0.9381 & 1.3316& 0.2562& 0.7444 \\ 
\hhline{|=|=|=|=|=|=|=|}
 Naive & 8.8172 &  0.8980 &6.9603 & 1.0548 & 278.4124& 1.0329 \\ 
 \hline
\end{tabular}
  \caption{Observability data, 1-minute resolution, context 512; normalized by Naive (except Naive)}
\label{tab:1min_overall}
\end{table}
\end{center}

\begin{center}
\begin{table}[ht]% Try here, and then top
\small
\centering
\renewcommand{\arraystretch}{1.2}
\begin{tabular}{ |l|c|c|c|c|c|c| } 
 \hline
  & MSE & MAE & MASE & sMAPE & MSIS & CRPS \\ 
\hhline{|=|=|=|=|=|=|=|}
 Cisco TSM (512, 512) & \textbf{0.9500} & \textbf{0.5011} & \textbf{0.5257} & \textbf{0.8010} & \textbf{0.2248} & \textbf{0.4381}  \\ 
\hline
 TimesFM-2.5 (1024) & 0.9606 & 0.5819  & 0.6159 & 0.8355 & 0.3142 & 0.5099\\ 
\hline
 TimesFM-2.0 (1024) & 1.0246 & 0.6234 & 0.7211 & 0.9313 & 0.3251 & 0.5416 \\ 
\hline
 Chronos-2 (1024) & 0.9647 &  0.5511& 0.6052 & 0.8333& 0.3078& 0.4831 \\ 
\hline
 Chronos-Bolt (1024) & 1.1681 & 0.7316  & 0.7717 &1.0263 & 0.3245 & 0.6383 \\ 
\hline
 Toto-1.0 (1024) &  0.9589& 0.5497  & 0.5733 & 0.8109& 0.3091 & 0.4821 \\ 
\hline
 AutoARIMA (1024) & 1.0669 & 0.8022 & 0.8167 & 1.2522 & 0.6256 & 1.010 \\ 

\hhline{|=|=|=|=|=|=|=|}

 Naive (1024) & 25.3064 & 0.8748 &11.7453 & 0.9802& 469.8138& 0.8787 \\ 
\hline
 \hline
  \hline
\end{tabular}
  \caption{Observability data, 1-minute resolution, context 1024; normalized by Naive (except Naive)}
\end{table}

\end{center}

\begin{center}
\begin{table}[!htb]% Try here, and then top
\centering
\small

\renewcommand{\arraystretch}{1.2}
\begin{tabular}{ |l|c|c|c|c|c|c| } 
 \hline
  & MSE & MAE & MASE & sMAPE & MSIS & CRPS \\ 
\hhline{|=|=|=|=|=|=|=|}
 Cisco TSM (512, 512) & \textbf{0.7307} & \textbf{0.5510} & \textbf{0.5439}& \textbf{0.6965} & \textbf{0.1582} & \textbf{0.4444} \\ 
\hline
  TimesFM-2.5 (512) & 0.7703	 &	0.6222 &	0.6713 &	0.7762	& 0.1883 & 0.5082 \\ 
\hline
 TimesFM-2.0 (512) & 0.8016 &  0.6593 & 0.7793 & 0.8247 & 0.2357 & 0.5358 \\ 
\hline
 Chronos-2 (512) &  0.7649 &  0.5984 & 0.6255 &  0.7291 & 0.1866 &  0.4870\\ 
\hline

 Chronos-Bolt (512) & 0.8052 & 0.6992 & 0.7403 & 0.8619 & 0.3150 & 0.5980 \\ 
\hline

 Toto-1.0 (512) & 0.7781  & 0.6164  & 0.6340 & 0.7388 & 0.2189 &  0.5039\\ 
\hline

 AutoARIMA (512) & 30.7992 & 0.8766 & 0.9541 & 1.2324 & 0.2922& 0.7428 \\ 
%\hline
\hhline{|=|=|=|=|=|=|=|}

 Naive & 4.1294 & 0.9336 & 5.4833 &1.0513 &219.3303 &  1.0001\\ 
 \hline

\end{tabular}
  \caption{Observability data, 5-minute resolution, context 512; normalized by Naive (except Naive)}
\end{table}
\end{center}

\begin{center}
\begin{table}[h]% Try here, and then top
\centering
\small

\renewcommand{\arraystretch}{1.2}
\begin{tabular}{ |l|c|c|c|c|c|c| } 
 \hline
  & MSE & MAE & MASE & sMAPE & MSIS & CRPS \\ 
\hhline{|=|=|=|=|=|=|=|}
 Cisco TSM (512, 512) & \textbf{0.9525} & \textbf{0.5482} & \textbf{0.5836} & \textbf{0.7134} & \textbf{0.3318} & \textbf{0.4642} \\ 
 \hline
 TimesFM-2.5 (1024) & 0.9590 &0.6060& 0.6335 & 0.7636&0.3526 & 0.5129\\ 
\hline
 TimesFM-2.0 (1024) & 0.9878&0.6381 & 0.7448&0.8155& 0.3667& 0.5354 \\ 
\hline
 Chronos-2 (1024) & 0.9629&0.5720&0.6245 &0.7290 & 0.3393 &  0.4840\\ 
 \hline

 Chronos-Bolt (1024) &1.0601 &0.6865&0.7607&0.8626& 0.3648& 0.5761 \\ 
\hline%\hhline{|=|=|=|=|=|=|=|}

 Toto-1.0 (1024) & 0.9572 & 0.5799  & 0.6286 & 0.7291& 0.3543&  0.4911\\ 
\hline
%\hhline{|=|=|=|=|=|=|=|}
 
 AutoARIMA (1024) & 3.1151 & 0.8603 & 0.8590 & 1.1783 & 0.5286 &  0.9566\\ 
\hhline{|=|=|=|=|=|=|=|}
 
 Naive (1024) & 20.3347 & 0.9174 &13.3537 & 0.9716& 534.1749& 0.8396 \\ 

 \hline
 
\end{tabular}
  \caption{Observability data, 5-minute resolution, context 1024; normalized by Naive (except Naive)}
\end{table}
\end{center}

\subsubsection{GIFT-Eval Benchmark} \label{gift_data}

To facilitate a fairer comparison among models, we remove from the GIFT-Eval benchmark those datasets known to be part of the TimesFM 2.0 training corpus (noted as ``non-leaking'' in the evaluation results). Regardless of the actual resolution of a time series, we supply a multiresolution context by aggregating non-overlapping segments of length 60 (of up to 30,720 data points preceding a horizon) and also providing up to 512 points preceding the horizon. Since these data are labeled with a seasonality, we report SeasonalNaive instead of Naive as another baseline, and use this to normalize the errors. We adopt the standard approach of normalizing errors per dataset before applying a shifted geometric mean.

We report performance separately on windows with at least 512 data points as context and those with fewer than 512. This evaluation demonstrates that the architectural and data modifications of our model did not ``destroy'' the capabilities of TimesFM 2.0. Indeed, compared to TimesFM 2.0, we obtain better performance on the observability datasets, and overall better performance on the long context subset of GIFT-Eval. We expect that similar modifications could be successfully applied to other TSFMs as well. Section \ref{qual} shows qualitatively, via examples, how the long context leads to better predictions.

\begin{center}
\begin{table}[ht]% Try here, and then top
\centering
\small
\renewcommand{\arraystretch}{1.2}
\begin{tabular}{ |l|c|c|c|c|c|c| } 
 \hline
  & MSE & MAE & MASE & sMAPE & MSIS & CRPS \\ 
\hhline{|=|=|=|=|=|=|=|}
 Cisco TSM (512, 512) & 0.5423 & 0.6980 & 0.7365 & 1.1053 & 0.5649 & 0.5508 \\ 
% \hhline{|=|=|=|=|=|=|=|}
\hline
 % TimesFM-2.5 (512) &  &  & & & &  \\ 
 % \hline
 % TimesFM-2.5 (512, 512) &  &  & & & &  \\ 
 % \hline
 TimesFM-2.5 (1024) & \textbf{0.5111} & 0.6635 & 0.6828 & \textbf{0.9416} & 0.5230	 & 0.5247 \\ 
% \hhline{|=|=|=|=|=|=|=|}
\hline
 
 TimesFM-2.0 (1024) & 0.5800   & 0.7119	 & 0.7620 & 1.1078	 & 0.6154 &0.5680\\ 
% \hhline{|=|=|=|=|=|=|=|}
 \hline

 Chronos-2 (1024) & 0.5263  & 0.6569	 & \textbf{0.6738}	 & 1.0395 & \textbf{0.4691}	 & 0.5122\\ 
% \hhline{|=|=|=|=|=|=|=|}
\hline
 
 Chronos-Bolt (1024) & 0.6149 & 0.7382 & 0.7916 & 1.1633 & 0.6158	 & 0.5928 \\ 
% \hhline{|=|=|=|=|=|=|=|}
\hline

 Toto-1.0 (1024) & 0.5292  & \textbf{0.6392} & 0.6784 & 0.9493 & 0.5080 & \textbf{0.5085} \\ 
\hline
 AutoARIMA (1024) & 0.7671 & 0.9115 & 0.9121 & 1.1169 & 0.8033	 & 0.8266 \\ 
\hhline{|=|=|=|=|=|=|=|}

 SeasonalNaive & 4895.5820 & 29.9572 & 1.1846& 0.4302&14.6630	&0.3990\\
   \hline

\end{tabular}
  \caption{Filtered (non-leaking) GIFT-Eval benchmark, context $\geq 512$, prediction length capped at 128; normalized by SeasonalNaive (except SeasonalNaive)}
\end{table}
\end{center}

\begin{center}
\begin{table}[ht]
\centering
\small
\renewcommand{\arraystretch}{1.2}
\begin{tabular}{ |l|c|c|c|c|c|c| } 
 \hline
  & MSE & MAE & MASE & sMAPE & MSIS & CRPS \\ 
\hhline{|=|=|=|=|=|=|=|}
 Cisco TSM (512, 512) & 0.6038 & 0.7399 & 0.7658 & 1.0967 & 0.5564 & 0.5783 \\ 

\hline

 TimesFM-2.5 (1024) & \textbf{0.5176} & 0.6701 &0.6914 & \textbf{0.9191} &0.5129 & 0.5209 \\ 

\hline

 TimesFM-2.0 (1024) & 0.5795 & 0.7156 & 0.7622 & 1.0743 & 0.5935 & 0.5601 \\ 

\hline
 
 Chronos-2 (1024) & 0.5247 & \textbf{0.6619} & \textbf{0.6808} &1.0140 & \textbf{0.4635}& \textbf{0.5084} \\ 

\hline
 
 Chronos-Bolt (1024) & 0.6120 & 0.7388 &0.7881 & 1.1209 & 0.5860 &  0.5837 \\ 

\hline
 
 Toto-1.0 (1024) & 0.5529 &0.6626  &0.7032 & 0.9463 &0.4955 &  0.5110\\ 
\hline
 AutoARIMA (1024) & 0.7416 &0.8909  &0.8969 & 1.0685 &0.7683 &  0.7914\\ 
\hhline{|=|=|=|=|=|=|=|}
 
 SeasonalNaive & 7955.5673	 & 41.3166 & 1.5610	 & 0.4289 & 20.3383 &  0.4410 \\ 
 \hline

 \hline
\end{tabular}
  \caption{Filtered (non-leaking) GIFT-Eval benchmark, prediction length capped at 128; normalized by SeasonalNaive (except SeasonalNaive)}
  \label{tab:filtered_eval}
\end{table}
\end{center}

\subsection{Qualitative Results} \label{qual}

To better understand how the 1-hour context influences forecasting performance, we compare model predictions across a spectrum of 1-hour context lengths and corresponding error levels.
More information from the 1-hour context usually improves forecasting accuracy as the model can learn the dynamics more reliably. However, certain highly volatile or irregular series remain challenging even with extended historical context, indicating that forecasting quality depends not only on context length but also on the intrinsic complexity of the signal.

\begin{figure}[H]
    \centering
    \includegraphics[width=0.9\textwidth, height=9cm]{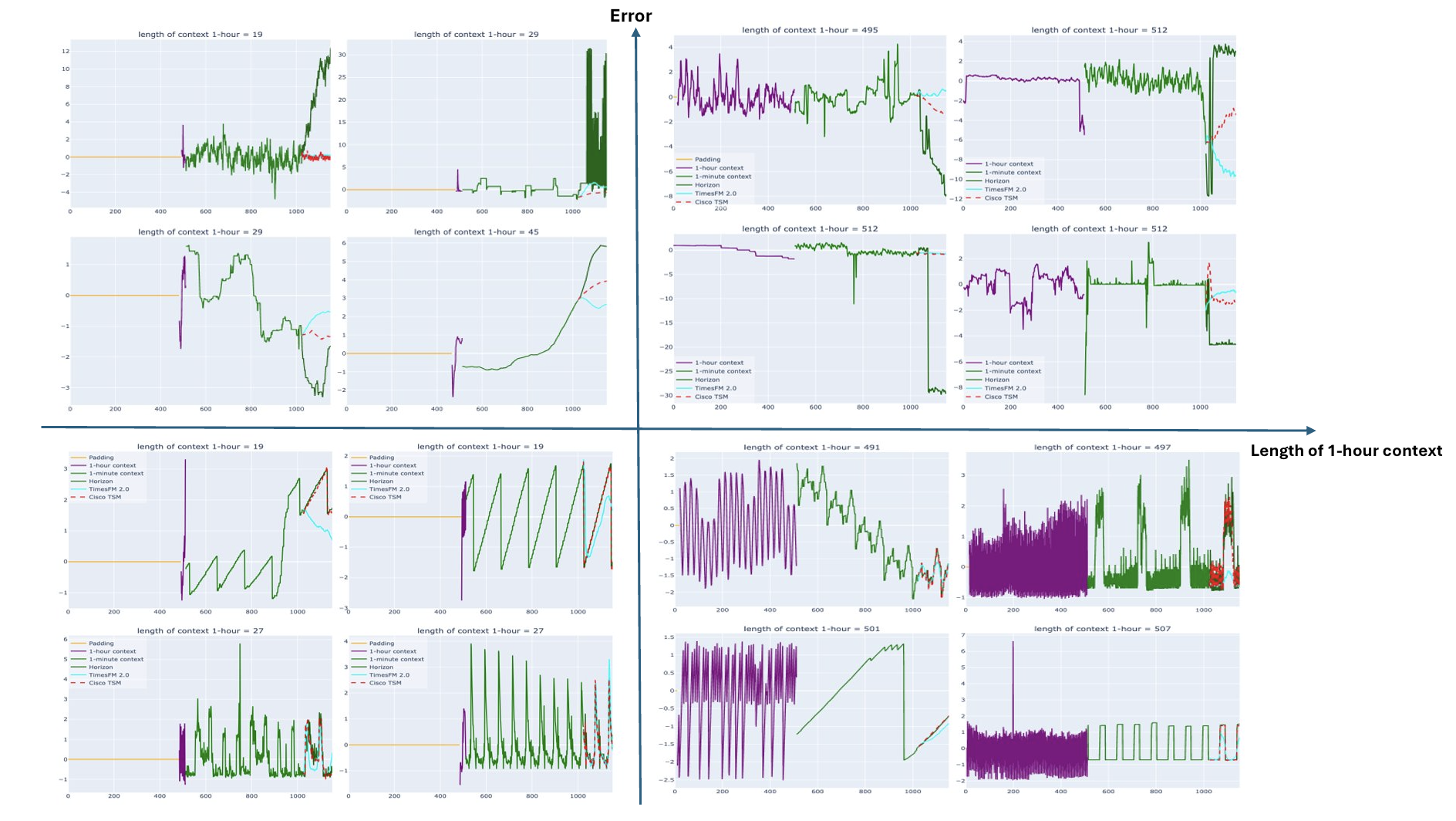}
    \caption{Comparing Short/Long 1-hour context vs.~Low/High Error}
    \label{fig:quadrants}
\end{figure}

In Figure \ref{fig:quadrants}, the high error quadrants suggest our model has trouble adapting to late-breaking changes. The bottom right quadrant shows examples where our model uses longer-term structure and/or simply ignores temporary noise, leading to better predictions. In the bottom left quadrant, our model appears to better understand the dynamics near the boundary.

\begin{figure}[t]
    \centering
    \includegraphics[width=0.7\textwidth, height=9cm]{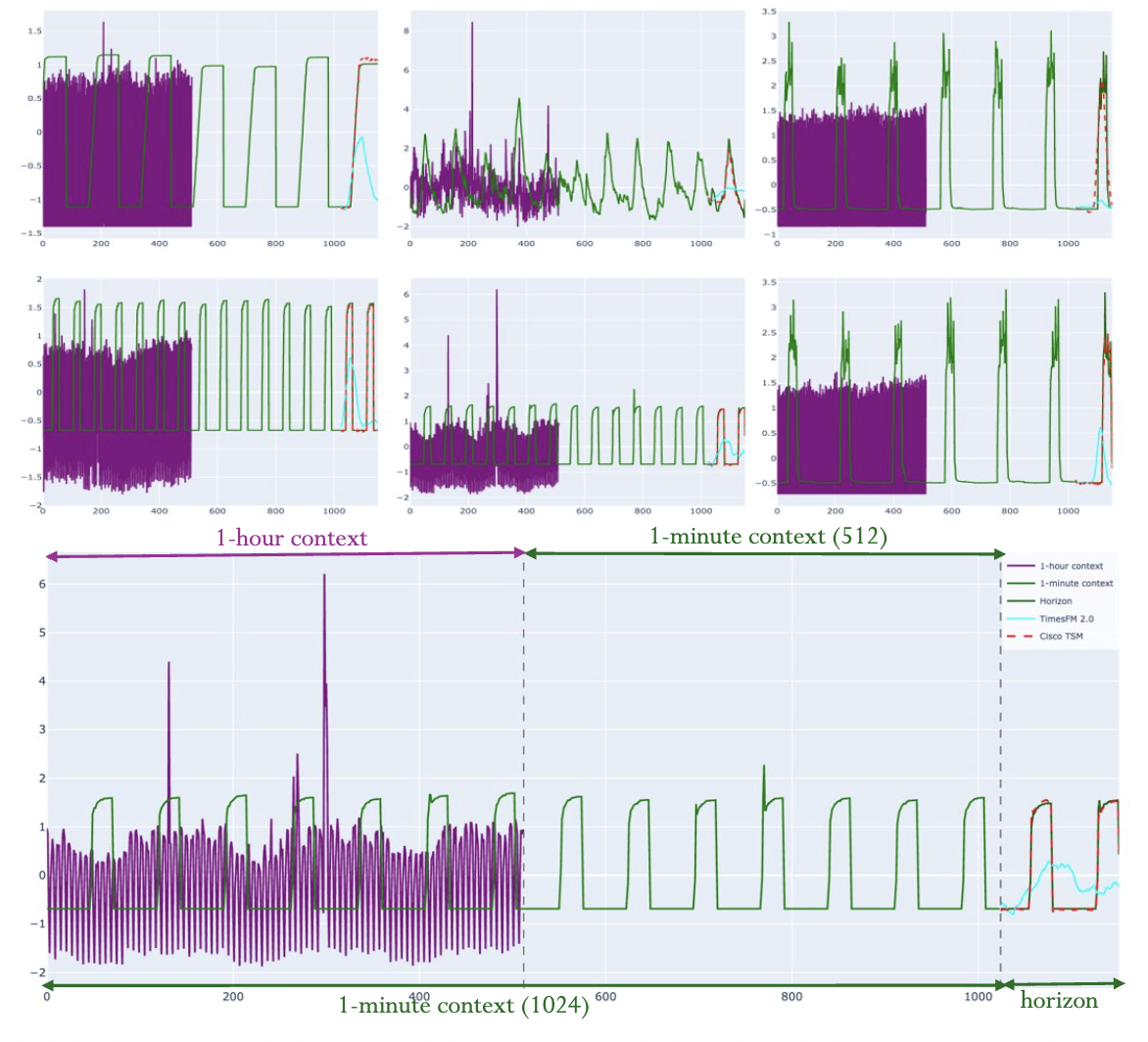}
    \caption{Examples where longer 1-hour context increases prediction accuracy for time series with strong periodic patterns and moderate noise; 1024-minute history also shown.}
    \label{fig:example_}
\end{figure}

\begin{figure}[H]
    \centering
    \includegraphics[width=0.7\textwidth, height=9cm]{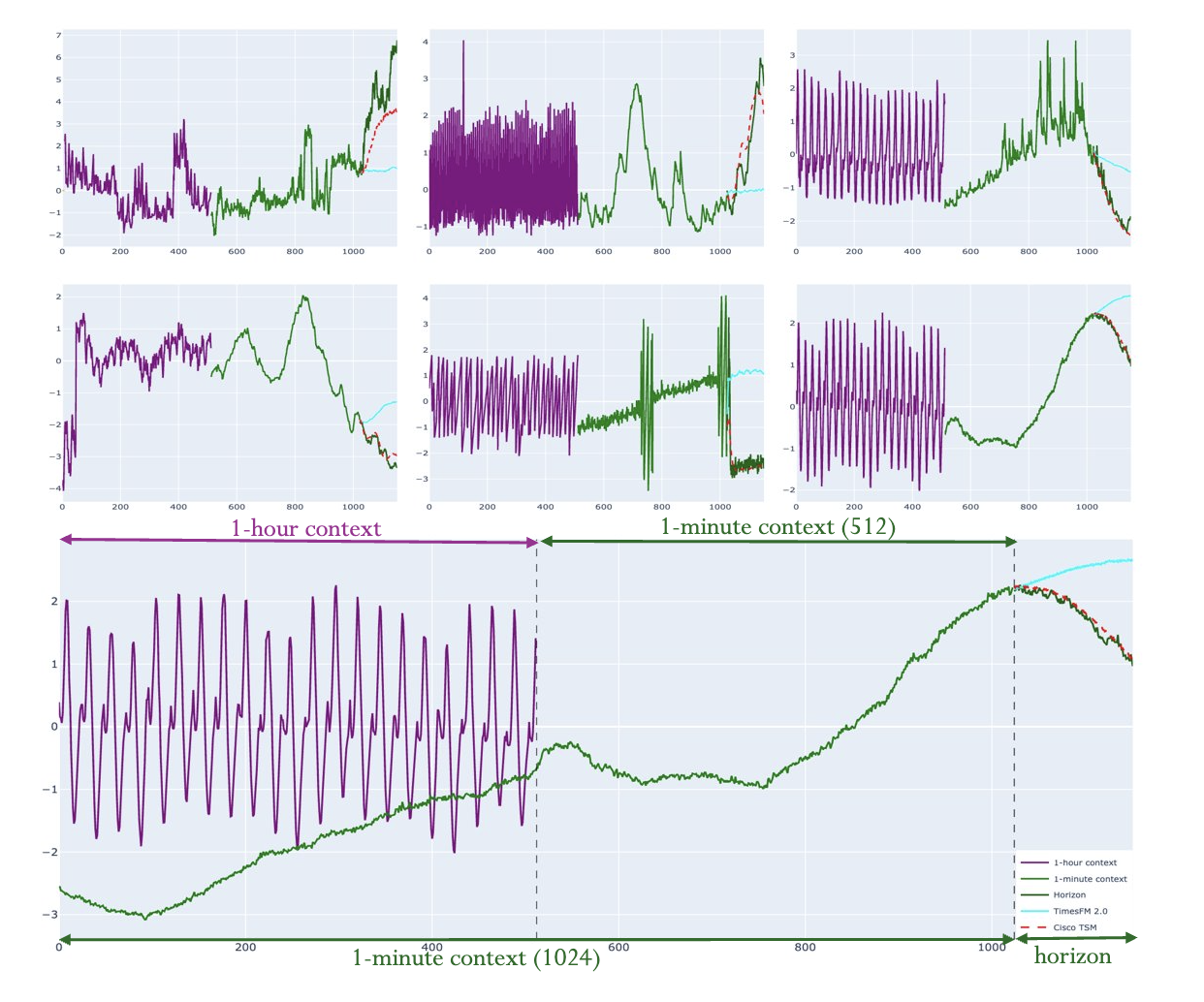}
    \caption{
    Examples where longer 1-hour context increases prediction accuracy for time series with longer-term periodicity and shorter-term trend possibly at odds with one another; 1024-minute history also shown.
    }
    \label{fig:example__}
\end{figure}
\begin{figure}[H]
    \centering
    \includegraphics[width=0.7\textwidth, height=9cm]{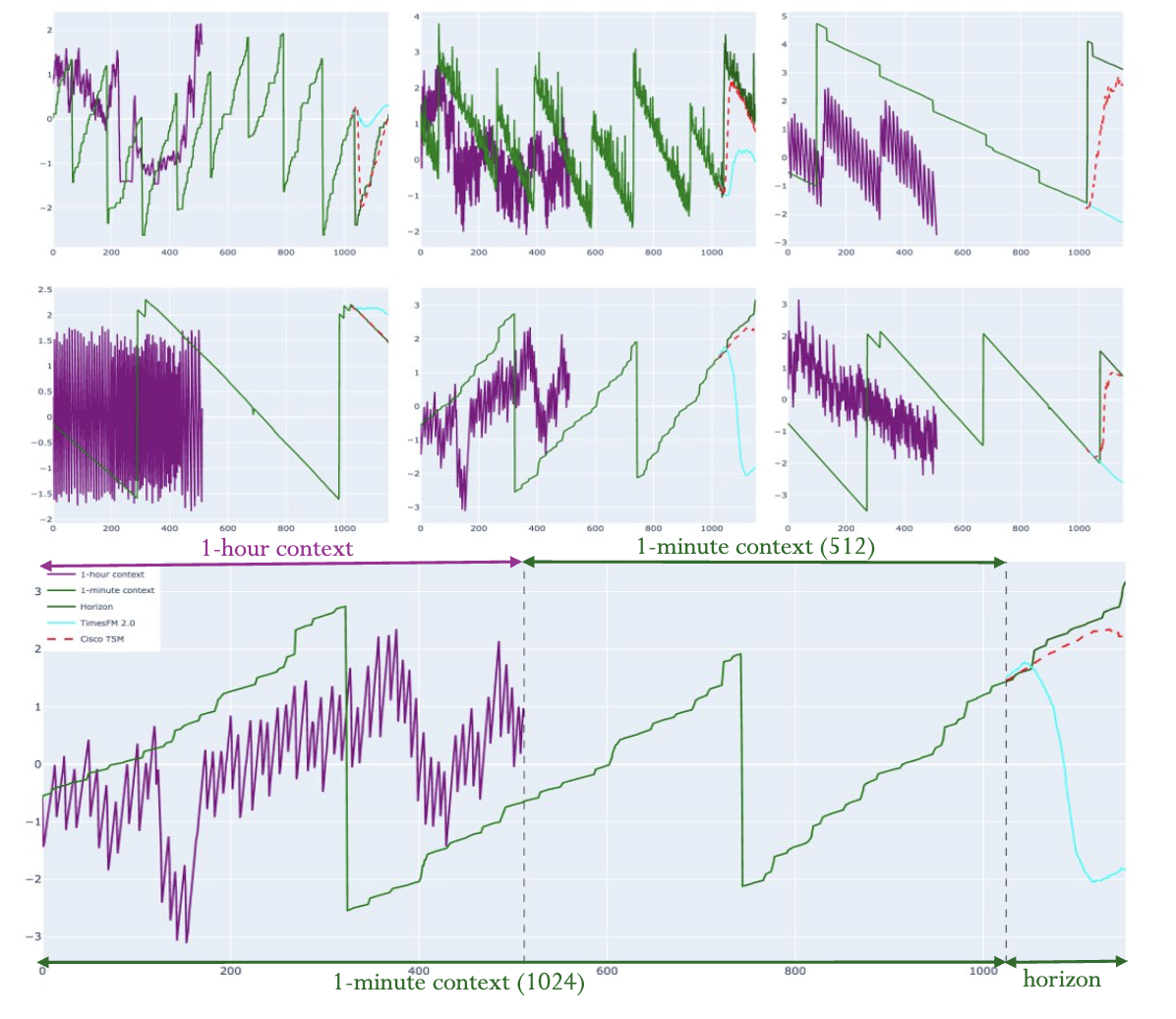}
    \caption{Examples where longer 1-hour context increases prediction accuracy for time series with sawtooth patterns by providing more dynamics, similar to Figure \ref{fig:first_sawtooth}; 1024-minute history also shown.}
    \label{fig:example___}
\end{figure}

\begin{figure}[H]
    \centering
    \includegraphics[width=0.7\textwidth, height=9cm]{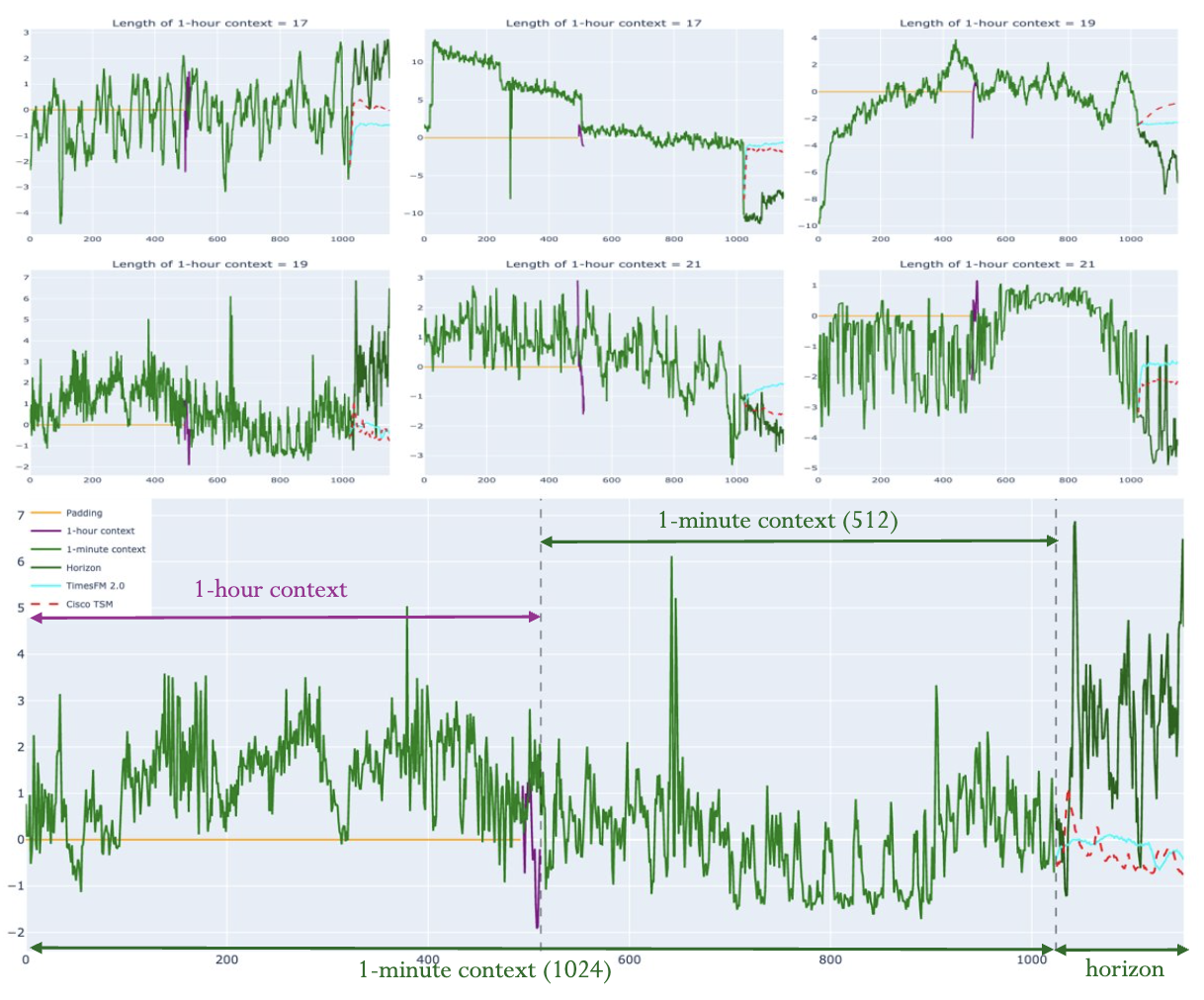}
    \caption{Examples where shorter 1-hour context and sharp transitions lead to poor predictions. Padding in 1-hour context is highlighted; 1024-minute history also shown.}

    \label{fig:example____}
\end{figure}

\subsection{Ablation Studies} \label{sec:ablation}

Our model is obtained by continued pre-training (CPT) on multiresolution data using both a special token and resolution embeddings to distinguish the two parts of the context. More natural than TimesFM 2.0 (512, 512) (i.e., simply treating a multiresolution context as though it were a single resolution) are the following configurations.

\begin{itemize}
    \item CPT on multiresolution data with no architectural modifications (CONCAT);
    \item CPT on multiresolution data using only the resolution embeddings (RE);
    \item CPT on multiresolution data using only the special token (ST);
    \item CPT on multiresolution data using both resolution embeddings and special token (RE+ST).
\end{itemize}

We carried out a series of ablation studies based on these approaches that guided the architectural choices of our model. For these experiments, we used two smaller training sets (10B, 35B in size) following the same distribution as mentioned in Section \ref{sec:diversity}. The same evaluation sets are used for all configurations.

\begin{table}[h]
    \centering
    \small
    \renewcommand{\arraystretch}{1.2}
    \begin{tabular}{|c|c|c|c|c|c|c|}
    \hline
    \multirow{2}{*}{Model} & \multicolumn{3}{c|}{GIFT-Eval} & \multicolumn{3}{c|}{Observability, 1-minute} \\
    \cline{2-7}
    & MAE & MASE & CRPS
    & MAE & MASE & CRPS \\
    \hhline{|=|=|=|=|=|=|=|}
    CPT (CONCAT) 
    & 0.7377 & 0.7984 & 0.6742 
    & \textbf{0.5144} & \textbf{0.4946} & \textbf{0.4451} \\
    \hline
    CPT (RE) 
    & 0.7379 & 0.7984 & 0.6737 
    & 0.5227 & 0.5010 & 0.4522 \\
    \hline
    CPT (ST) 
    & 1.0262 & 1.0179  & 0.9360 
    & 1.3361 & 1.0412  & 1.2966 \\
    \hline
    CPT (RE+ST) 
    & \textbf{0.7348} & \textbf{0.7890} & \textbf{0.6718} 
    & 0.5246 & 0.4986 & 0.4552 \\
    \hhline{|=|=|=|=|=|=|=|}

    Naive 
    & 1.4436 & 12.4395 & 1.0166
    & 0.8980 & 6.9603 & 1.0329 \\
    \hline
    
    \end{tabular}
    \caption{Performance of experiments conducted on a 10B scale training dataset. Each row is normalized by Naive (except for Naive).}
    \label{tab:ablation-10b}
\end{table}

\begin{table}[h]
    \centering
    \small
    \renewcommand{\arraystretch}{1.2}
    \begin{tabular}{|c|c|c|c|c|c|c|c|}
    \hline
    \multirow{2}{*}{Model} & \multicolumn{3}{c|}{GIFT-Eval} & \multicolumn{3}{c|}{Observability, 1-minute} \\
    \cline{2-7}
    & MAE & MASE & CRPS
    & MAE & MASE & CRPS \\
    \hhline{|=|=|=|=|=|=|=|}
    CPT (CONCAT) 
    & 0.7311 & 0.7815 & 0.6676 
    & 0.5022 & 0.4803 & 0.4336 \\
    \hline
    CPT (RE) 
    & 0.7285 & \textbf{0.7660} & 0.6656 
    & 0.5177 & 0.4910 & 0.4486 \\
    \hline
    CPT (ST) 
    & \textbf{0.7263} & 0.7678 & \textbf{0.6634} 
    & \textbf{0.4994} & \textbf{0.4767} & \textbf{0.4310} \\
    \hline
    CPT (RE+ST) 
    & 0.7299 & 0.7835 & 0.6670 
    & 0.5018 & 0.4816 & 0.4332 \\
    \hhline{|=|=|=|=|=|=|=|}

    Naive & \multicolumn{6}{c|}{(same as Table \ref{tab:ablation-10b})} \\
    \hline
    
    \end{tabular}
    \caption{Performance of experiments conducted on a 35B scale training dataset. Each row is normalized by Naive (except for Naive).}
    \label{tab:ablation-35b}
\end{table}

The experiments conducted with only special token (ST) or only resolution embeddings (RE) exhibited numerical inconsistencies that we could not explain. Their parameters constitute only a tiny fraction of the total number of parameters of the model, but the model's performance changes drastically with the scale of the data (Tables \ref{tab:ablation-10b} and \ref{tab:ablation-35b}) and was inconclusive.

Of the other two approaches, we found that our approach of combining both ST and RE yielded either similar or better performance than CPT (CONCAT). We also found that our approach (RE+ST) generally converges faster than does CPT (CONCAT).

\section{Conclusion}
The Cisco Time Series Model is a univariate zero-shot forecaster based on the transformer. It introduces a novel multiresolution architectural pattern that enables efficient use of long context, particularly important for operational scenarios where detail and timeliness must be balanced against accuracy and incorporation of longer-term structures.
It displays superior performance on observability datasets while retaining solid performance on a general benchmark. Altogether, this provides a path to incorporating both data from new domains and architecture modifications to existing time series foundation models, which not only advances the state of the art in time series foundation modeling but also results in increased forecasting accuracy for time series data known to be of high business value.

\subsection*{Acknowledgments}
It is our pleasure to thank Shruti Sharma,  Vishwambhar Upadhyay, and Meet Vaddoriya for data engineering support; Park Kittipatkul and Mike Kruze for helping us understand the observability system; Arijit Mukherji for many interesting discussions and ideas; Brian Goleno, Nick Ma, and Sonal Pardeshi for endless advocacy and guiding us towards the right business outcomes; and Matthew McCloskey for tireless organizational support.

\printbibliography

\end{document}